\documentclass[sigconf]{acmart}
\usepackage{colortbl}
\usepackage{caption}
\usepackage{subcaption}

\AtBeginDocument{%
  \providecommand\BibTeX{{%
    \normalfont B\kern-0.5em{\scshape i\kern-0.25em b}\kern-0.8em\TeX}}}

\setcopyright{acmcopyright}
\copyrightyear{2024}
\acmYear{2024}
\acmDOI{XXXXXXX.XXXXXXX}

\acmConference[RLC @ WSDM '24]{the 17th ACM International
Conference on Web Search and Data Mining}{March 04--08,
  2024}{Merida, Mexico}
\acmPrice{15.00}
\acmISBN{978-1-4503-XXXX-X/18/06}

\begin{document}

\title{Unveiling the Potential of BERTopic for Multilingual Fake News Analysis - Use Case: Covid-19}

\author{Karla Schäfer}\orcid{0009-0004-1731-7925}
\affiliation{%
  \institution{Fraunhofer SIT | ATHENE}
  \streetaddress{Rheinstr.75}
  \city{Darmstadt}
  \country{Germany}
  \postcode{64295}
}
\email{karla.schaefer@sit.fraunhofer.de}

\author{Jeong-Eun Choi}\orcid{0009-0007-3599-0808}
\affiliation{%
  \institution{Fraunhofer SIT | ATHENE}
  \streetaddress{Rheinstr.75}
  \city{Darmstadt}
  \country{Germany}
  \postcode{64295}
}
\email{jeong-eun.choi@sit.fraunhofer.de}

\author{Inna Vogel}
\affiliation{%
  \institution{Fraunhofer SIT | ATHENE}
  \streetaddress{Rheinstr.75}
  \city{Darmstadt}
  \country{Germany}
  \postcode{64295}
}
\email{inna.vogel@sit.fraunhofer.de}

\author{Martin Steinebach}
\affiliation{%
  \institution{Fraunhofer SIT | ATHENE}
  \streetaddress{Rheinstr.75}
  \city{Darmstadt}
  \country{Germany}
  \postcode{64295}
}
\email{martin.steinebach@sit.fraunhofer.de}




\begin{abstract}
Topic modeling is frequently being used for analysing large text corpora such as news articles or social media data. 
BERTopic, consisting of sentence embedding, dimension reduction, clustering, and topic extraction, is the newest and currently the SOTA topic modeling method. However, current topic modeling methods have room for improvement because, as unsupervised methods, they require careful tuning and selection of hyperparameters, e.g., for dimension reduction and clustering.
This paper aims to analyse the technical application of BERTopic in practice. For this purpose, it compares and selects different methods and hyperparameters for each stage of BERTopic through density based clustering validation and six different topic coherence measures.
Moreover, it also aims to analyse the results of topic modeling on real world data as a use case. For this purpose, the German fake news dataset (GermanFakeNCovid) on Covid-19 was created by us and in order to experiment with topic modeling in a multilingual (English and German) setting combined with the FakeCovid dataset. With the final results, we were able to determine thematic similarities between the United States and Germany. Whereas, distinguishing the topics of fake news from India proved to be more challenging.
\end{abstract}



\maketitle

\section{Introduction}
BERTopic is used in various publications for extracting topics of text corpora \cite{9631322,ABUZAYED2021191,comparison}.
For BERTopic, the steps of sentence embedding, dimension reduction, clustering, and topic extraction require a careful selection of different hyperparameters. However, there is no optimal solution for the selection of these hyperparameters. Metrics such as density based clustering validation (DBCV) and topic coherence can be used to evaluate the resulting clustering or topic representation.
However, the best hyperparameters determined through these evaluations do not always result in the best representation of the topics.

In this paper, we analyse and compare two different embeddings and evaluate the results of DBCV across different hyperparameter settings for UMAP, which is used for dimension reduction, and HDBSCAN, which is used for clustering. Subsequently, the results of c-TF-IDF and KeyBERT as topic extraction methods will be evaluated using six different topic coherence measures.

To observe the performance of topic modeling in a practical application, we decided to analyse data containing fake news published during the Covid-19 pandemic. During the pandemic, huge amounts of information were generated worldwide, interspersed with false, falsified or misleading information. Topic modeling can be an effective tool to gain an overview of vast amount of data and this can be used to counteract against the spread of false information. For example, fact-checkers are confronted with a timely decision on claims that are relevant and worthy of verification. The identified topics on which fake news has frequently been disseminated can serve as an initial indication for the selection of such check-worthy claims. Despite its potential, there had been less research on how topic modeling could be used for the fact-checking process in practical scenes or how it can be used to bring more insight into fake news in general.
For this purpose, a new dataset with German fake news on the Covid-19 pandemic was created by us. With an additional dataset (FakeCovid), the topics of fake news on Covid-19 from three different nations (India, United States and Germany) in two different languages (English and German) were analysed using different settings of BERTopic. The final topics identified are then compared to each other for determining thematic similarities.

The paper is structured as follows:
Section \ref{related_work} presents the related work.
In Section \ref{datasets}, the examined datasets are introduced in detail.
First, the new dataset of Covid-19 related fake news from Germany is presented, followed by the FakeCovid dataset of \citet{shahifakecovid}.
Section \ref{experimental_methodology} discusses the experimental methodology which consists of data preprocessing, sentence embedding, dimension reduction, clustering, and topic extraction.
Subsequently, the results are presented and discussed in Section \ref{results_analysis} and \ref{sec:discssion}.
The paper finishes with a conclusion and future work (Section \ref{conclusion}).

\section{Related Work}\label{related_work}
The related work briefly discusses 1) different topic modeling techniques, 2) dimension reduction and clustering algorithms, and 3) topic modeling on Covid-19 news in particular. 

\subsection{Topic Modeling Methods}\label{review_topicmodeling_methods}
LDA and NMF are two of the most popular methods for topic modeling \cite{ABUZAYED2021191, comparison}.
However, the text representations of both models are too simple so that most semantic and syntactic features are lost.
In addition, the number of topics to be identified must be known in advance.
Transformer-based language models such as BERT have achieved excellent results in the field of Natural Language Understanding.
Because of these successes, pre-trained language models have been applied to topic modeling, for example in Top2Vec or BERTopic.

\citet{comparison} compared LDA, NMF, Top2Vec and BERTopic in an analysis of X (formerly Twitter) posts about travel and the Covid-19 pandemic. 
The observed dataset includes 31,800 tweets collected by searching for the terms \#covidtravel and the combination of \#covid and \#travel.
The results of LDA and NMF as well as those of Top2Vec and BERTopic were compared.
Overall, BERTopic and NMF provided the best results, with clear differences in the topics identified, and they proposed that BERTopic was found to be the most promising.
Similarly, \citet{9631322} and \citet{ABUZAYED2021191} found that BERTopic produced results with higher topic diversity and quality and higher Normalized Pointwise Mutual Information (NPMI), than LDA.

\subsection{Dimension Reduction and Clustering}

Topic modeling identifies topics by grouping items into similar topics.
Similar items can be determined by their representation in vector space.
For example, BERTopic uses a sentence transformer, where the documents are mapped to a 768 dimensional dense vector space.
As the number of dimensions increases, a meaningful distinction between similar and dissimilar objects becomes increasingly difficult.
In particular, assessing similarity on the basis of distance is no longer a meaningful criterion \cite{assent2012clustering}.
Although some sentence transformer models, such as all-mpnet-base-v2, have been designed to preserve the usefulness of euclidean distances, dimensionality reduction before clustering is recommended because most clustering algorithms cannot group high-dimensional inputs well \cite{sbert, 9640295}.
Well-known methods for dimension reduction are PCA \cite{pca}, t-SNE \cite{tsne} and UMAP (Uniform Manifold Approximation and Projection) \cite{umap}. UMAP was used in this paper for dimension reduction as it preserves more of the global structure with better runtime performance \cite{umap}.

One of the most popular clustering algorithms is K-Means \cite{perez2019k,ikotun2023k}, whereby the number of clusters in a given dataset must be specified beforehand.
Other methods apply density-based clustering which can identify densely populated regions of arbitrary shapes. The identified clusters are then positioned in high-density regions and outliers appear in low-density regions \cite{chen2019knn}.
Density-based clustering is one of the most widely used clustering methods, with density-based spatial clustering of applications with noise (DBSCAN) \cite{ester1996density} being the most well-known \cite{wu2023vizoptics}. 
Closely related to DBSCAN is OPTICS (Ordering Points To Identify the Clustering Structure) \cite{ankerst1999optics} and HDBSCAN \cite{campello2013density}, the hierarchical version of DBSCAN. Both can identify clusters with varying densities and are therefore more appropriate for large datasets.

\subsection{Topic Modeling on Covid-19 News}
Topic modeling has already been applied to Covid-19 news for various purposes.
The study by \citet{topicmodelingvisualization} seeks to determine the public's concern and experience with the Covid-19 pandemic by analysing data from social media, more specifically X.
The main topics of tweets in English on Covid-19 were identified with LDA.
\citet{concernstweeterscovid19} analysed messages from X, with the aim of identifying the most important topics posted in relation to the Covid-19 pandemic.
For this purpose, a dataset was created containing tweets in English identified with the search terms 'corona', '2019-nCov' and 'COVID-19'.
Twelve topics were determined with LDA, which were grouped into four main topics: 1) origin of the virus, 2) its sources, 3) its impact on people, countries and the economy, and 4) ways of mitigating the risk of infection.
\citet{healthcommunicationcovid19} identified the patterns of media-driven health communication and the role of the media in the Covid-19 crisis in China.
For this, the WiseSearch database was used to collect news articles from China about Covid-19, with LDA 20 topics were identified and the three most popular topics were: 1) prevention and control methods, 2) medical treatment and research, and 3) global or local social and economic influences.

In addition to identifying all topics related to Covid-19, work has also been published on identifying sub-topics related to a specific main topic in context of the Covid-19 pandemic.
For instance, \citet{yin2022sentiment} and \citet{MELTON20211505} have examined contributions specifically to vaccination against Covid-19 using LDA. 
Others have focused on comparing topics between different countries. For example, \citet{investigating4nations} examined the topics and content of Covid-19 news from four countries. They created a database of more than 100,000 Covid-19 headlines and articles.
The articles were drawn from English-language websites of eight major newspapers from four countries (UK, India, Japan and South Korea).
Using Top2Vec for topic modeling they discovered the most frequently mentioned topics such as: 1) education, 2) business, 3) the USA and 4) sport.

Comparably little work has been done on applying topic modeling on fake news in particular.
\citet{9457813} analysed a dataset of fake news in general, with texts in Brazilian Portuguese (Fake.Br corpus).
Through LDA, the 10 most descriptive words per topic of fake and real news were identified, but without merging them to topic labels.
In general, fake news is studied in context of detection methods but is less discussed in terms of topic modeling.

Numerous studies have been carried out on the identification of topics in Covid-19 news.
However, most of the work used LDA, despite the good results of BERTopic and Top2Vec.
Therefore, to the best of our knowledge, this work is the first to use transformer-based language models to analyse fake news of the Covid-19 pandemic and applying clustering for finding (dis-)similarities between the topics in the fake news of different nations.
Moreover, it is the first study to look at fake news from three different countries in two different languages.

\section{Datasets}\label{datasets}
GermanFakeNCovid, consists of German-language fake news on Covid-19. 
It was created and published by us.\footnote{\url{https://doi.org/10.5281/zenodo.10958359}}
The second is the FakeCovid dataset published by \citet{shahifakecovid}.

\subsection{GermanFakeNCovid}
To create the GermanFakeNCovid dataset, the list of fact-checked news articles mentioned in the fact-checking organisations' reference source was first evaluated and articles related to Covid-19 were selected.
Further articles from various online sources such as Wochenblick, Facebook and online blogs like journalistenwatch.com were added to create a diverse source base.
Following the judgement of various fact-checking websites\footnote{Websites used for fact-checking: \url{https://dpa-factchecking.com/},\url{https://correctiv.org/faktencheck/},\url{https://faktencheck.afp.com/},\url{https://www.presseportal.de/},\url{https://www.tagesschau.de/faktenfinder/}.}, two annotators (both native German speaker) selected the articles identified as fake and used them to form the new dataset.
In total, the dataset contains 675 fake news articles published between January 2020 and February 2022, with the majority appearing in 2020 (288 articles) and 2021 (379 articles).

\subsection{FakeCovid}\label{fakecovid}
In this analysis, we aim to analyse fake news from different countries in two different languages. 
\citet{shahifakecovid} proposed FakeCovid, the first multilingual dataset of 7623 fact-checked news articles for Covid-19, collected from January 4th, 2020 to July 1th, 2020.
Fact-checked articles were collected and the truthfulness rate of each article were calculated, following the judgment of 92 different fact-checking websites.
This dataset, which includes posts in 40 languages from 105 countries, was also used as a basis for CheckThat! Task 3 in CLEF2021 \citep{Murayama2021DatasetOF}.
Based on \citet{Dulizia2021FakeND}'s analysis, the FakeCovid dataset is the only multilingual dataset to date with a thematic focus on Covid-19.
\citet{Murayama2021DatasetOF} examined 118 different publicly available datasets, including those related to fake news detection.
FakeCovid was the only dataset with news articles in more than one language and the only one in the news article field that dealt with Covid-19.
Additionally, FakeCovid contains fake news articles in one language but from different countries.

Unfortunately, this dataset only contains news articles about the Covid-19 pandemic from the first six months of 2020, while the GermanFakeNCovid dataset contains news from several years (2020-2022).
This may lead to a greater variety of German topics.
One possibility would be to use only those articles from the German dataset that were published within the six-month period.
However, with 83 articles, this would be too small for a thematic analysis.
Despite possible limitations caused by the time variance between the two datasets, because of similar numbers of articles available per country, the FakeCovid dataset was used.
India and the United States are represented with the most articles in the FakeCovid dataset and contain similar numbers of articles as the German dataset with 675 articles.

FakeCovid contains news articles in different languages per country. 
In order to obtain the topic-describing words in one language at a time during topic modeling, we decided to study each country in one language only.
For each country, the most represented language was selected and only these articles were used for the analysis. 
Furthermore, the dataset contains both fake and real news.
Since we are only interested in fake news topics, only the fake articles were considered.
Table \ref{tab:India_us_spain_dataset} provides an overview of the dataset splits used in the subsequent analysis.

\begin{table}[th]
\caption{Overview of dataset splits used for topic modeling.}
  \label{tab:India_us_spain_dataset}
  \scalebox{0.8}{
  \begin{tabular}{l|l|l|p{2.8cm}}
  \hline
  Dataset\cellcolor[gray]{0.9} & Nation\cellcolor[gray]{0.9} & Language\cellcolor[gray]{0.9} & Number of articles \cellcolor[gray]{0.9}\\
  \hline
  \hline
  FakeCovid & India & English & 653\\
  \hline
  FakeCovid & United States & English & 795\\
  \hline
  GermanFakeNCovid & Germany & German & 675 \\
  \end{tabular}
  }
 
\end{table}

\section{Experimental Methodology}\label{experimental_methodology}
\begin{figure*}[h!]
  \centering
  \includegraphics[width=\textwidth]{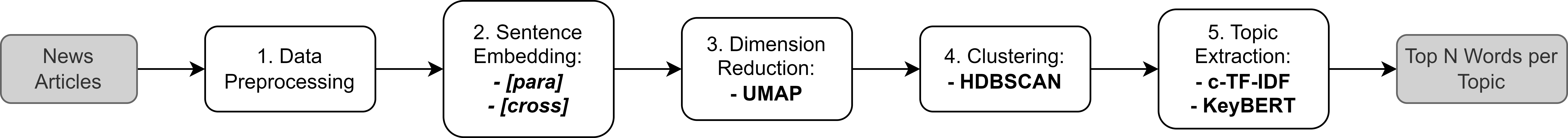}
  \caption{Flowchart of the topic modeling process. In bold the methods applied on each step.}
  \Description{First, the news articles are preprocessed, then sentence embedding takes place, followed by dimension reduction and clustering. The topics are then extracted from the clusters, which, with N words per topic, represents the result of the process.}
  \label{fig:process_topic_modeling}
\end{figure*}

To identify the topics of fake news, the datasets were first preprocessed before clustering was applied.
The structure of BERTopic, a recent framework for topic modeling consisting of four steps (sentence embedding, dimension reduction, clustering and topic extraction) \cite{grootendorst2022bertopic}, was used. 
As the clustering depends heavily on the sentence embedding, two different embedding methods were used and compared.
For analysing the dimension reduction and clustering different hyperparameter settings for UMAP and HDBSCAN were used and evaluated with DBCV. Two different topic extraction methods were applied and evaluated using six different topic coherence measures.
The entire process is summarised in Figure \ref{fig:process_topic_modeling}.

\subsection{Data Preprocessing}
Since we use pre-trained transformer models for sentence embedding, the texts can be passed to the model as they are.
However, since the data contains special characters, certain expressions are removed or replaced during preprocessing.
First, the emojis contained in the German dataset were removed.
Subsequently, the term Covid-19 was presented in a generalised way ('covid19'), as it occurs frequently in the articles and is of utmost importance for our task.

\subsection{Sentence Embedding}
In sentence embedding, the preprocessed texts were converted into dense vectors using transformer-based models from HuggingFace's sentence transformer package.
This package is based on Sentence BERT (SBERT), a work by \citet{sbert}.
We performed experiments with two different sentence embedding models and compared the results. First, we applied the commonly used \textit{paraphrase-multilingual-mpnet-base-v2}\footnote{\url{https://huggingface.co/sentence-transformers/paraphrase-multilingual-mpnet-base-v2}} model introduced by \citet{sbert}, called \textit{[para]} from now on in this paper. Secondly, we used the \textit{T-Systems-onsite/cross-en-de-roberta-sentence-transformer'}\footnote{\url{https://huggingface.co/T-Systems-onsite/cross-en-de-roberta-sentence-transformer}} model trained for German and English texts, called \textit{[cross]} from now on.
\textit{[cross]} provided slightly better results in terms of spearman's rank correlation between the cosine-similarity of the sentence embeddings and STSbenchmark labels on German data (0.8550), than the commonly used multilingual model \textit{[para]} (0.8355).
In English, \textit{[cross]} performed slightly worse with a correlation of 0.8660 (\textit{[para]} achieved 0.8682).
Because of these results and to evaluate the effect of embedding models trained on selected languages compared to multilingual models on  clustering results, we applied both models and compared the results.

\subsection{Dimension Reduction and Clustering}
As discussed in related work, UMAP \cite{umap} was used for dimension reduction.
For clustering, hierarchical DBSCAN (HDBSCAN) \cite{campello2013density} was used.
In UMAP, the parameters `n\_neighbors', `n\_components', `min\_dist' and `metric' can be set by the user.
In HDBSCAN three parameters have to be set: `min\_cluster\_size', which specifies the smallest number of points that the algorithm considers a cluster, the `cluster\_selection\_method' and `min\_samples'.
For finding the best hyperparameter setting for UMAP and HDBSCAN density based clustering validation (DBCV) \cite{moulavi2014density} was applied.

\subsection{Topic Extraction}
Two different topic extraction methods were used and compared.
For the first extraction method, c-TF-IDF based on TF-IDF (Term Frequency-Inverse Document Frequency)
, all documents in a cluster were merged so that each topic is represented by one large document. 
Then, the words of the documents were weighted using TF-IDF, a weighting scheme that reflects how important a word is to a document in a corpus. 
Before determining the TF-IDF weights, the stop words were removed from the documents using the NLTK Python library\footnote{\url{https://www.nltk.org/}}.
Then, the most relevant words per subject were selected to describe the topic.
The second extraction method, KeyBERT \cite{grootendorst2020keybert}, is initially a keyword extraction technique using sentence embeddings but is also frequently used to extract topics for BERTopic.

For evaluating the extracted topics, i.e. the words describing the topic, six evaluation measures for calculating the topic coherence were applied.
First, the coherence measures $c_v$, $u_{mass}$, $c_{uci}$ and $c_{npmi}$ were calculated using the gensim library\footnote{\url{https://radimrehurek.com/gensim/models/coherencemodel.html}} and its default settings. 
Secondly, the contextualized Topic Coherence Metrics (CTC) presented by \citet{rahimi2023contextualized} were used. 
CTC uses, similar to the work by \citet{stammbach2023revisiting}, large language models to evaluate the topic coherence. 
We applied semi-automated CTC, which outperformed the baseline metrics on short-text datasets \cite{rahimi2023contextualized}.
In semi-automated CTC, two scores were calculated using ChatGPT, namely $CTC_Intrusion$ and $CTC_Rating$.
$CTC_Intrusion$ is based on the topic words intrusion task, studied by \citet{chang2009reading}.
In this task, the topic coherence is assessed by identifying a coherent latent category for each topic and discovering the words that do not belong to that category. Instead of human subjects, \citet{rahimi2023contextualized} proposed to use ChatGPT to find a word describing the topic and finding the intruder words in the list of words examined.
For calculating $CTC_Rating$, ChatGPT was given the words describing the topic and the model was asked to rate the usefulness of the topic words for retrieving documents on a given topic on a 3-point scale where 3=“meaningful and highly coherent” and 0=“useless”.

\section{Results}\label{results_analysis}
The following presents the results of clustering with different hyperparameter settings and the evaluation of extracted topics using coherence measures.

\subsection{Clustering}
The DBCV score was used for finding the best hyperparameter settings of UMAP and HDBSCAN.
In Table \ref{tab:hyperparameter_DBCV}, first, the top results based on DBCV are presented.
For Germany, using \textit{[cross]} as embedding, three topics (and one outlier) were formed, with \textit{[para]} embedding five topics (and one outlier) were clustered.
For the United States, both embeddings and all top hyperparameter settings resulted in two clusters (no outliers). 
For the Indian fake news, two topics were formed. Using the \textit{[para]} embedding, additionally one cluster of outliers was created.
The best DBCV values were obtained for Germany and the United States on the \textit{[para]} embedding, for India on the \textit{[cross]} embedding.

Secondly, the hyperparameter settings of UMAP and HDBSCAN with the best DBCV score but also building more than two (or three, with outliers) clusters are given in Table \ref{tab:hyperparameter_DBCV}. With this, we also wanted to view more diverse clustering results and extract smaller topics, as the top hyperparameter setting only built two clusters for India and the United States. We assumed that with a low number of clusters, only the main topic of 'Covid-19 pandemic' may be detected in the fake news. Therefore, we decided to look at the results of a more diverse clustering as well. The analysis on topic extraction section shows that these two big clusters are hard to be recognized as topics.

\begin{table}[th]
    \caption{Overview of the parameter settings for UMAP and HDBSCAN, with 1) top results based on DBCV (Top) and 2) best DBCV scores with more than 2 (3 with outlier) clusters (\#C). All results displayed were obtained with cluster\_selection\_method=eom. Nation: Germany [G], United States [US], India [I]; O: One outlier cluster}
  \label{tab:hyperparameter_DBCV}
  \scriptsize
  \begin{tabular}{l|l|l|l|l|l|l|l}
  \hline
   \cellcolor[gray]{0.9} & \multicolumn{3}{c|}{UMAP \cellcolor[gray]{0.9}} & \multicolumn{2}{|c|}{HDBSCAN\cellcolor[gray]{0.9}}& \multicolumn{2}{|c}{Results\cellcolor[gray]{0.9}} \\
   \hline
    Nation;\cellcolor[gray]{0.9} & n\_nei- \cellcolor[gray]{0.9}& min\_\cellcolor[gray]{0.9} &  n\_com-\cellcolor[gray]{0.9} & min\_s-\cellcolor[gray]{0.9} & min\_clus-\cellcolor[gray]{0.9} & DBCV\cellcolor[gray]{0.9} & \# clus-\cellcolor[gray]{0.9}\\
   Embedding\cellcolor[gray]{0.9} & ghbors\cellcolor[gray]{0.9} & dist\cellcolor[gray]{0.9} & ponents\cellcolor[gray]{0.9} & amples\cellcolor[gray]{0.9} & ter\_size\cellcolor[gray]{0.9}  & score\cellcolor[gray]{0.9} & ter\cellcolor[gray]{0.9}\\
  \hline
  \hline
  Top- G \textit{[cross]} & 5 & 0.00 & 100 & 5 & 15 &  0.57 & 4 (O)\\
  Top- G \textit{[para]} &  5 & 0.09 & 20 & 10 & 30 & 0.6 & 6 (O)\\
\#C- G \textit{[cross]} & 50 & 0.09 & 2 & 5 & 20 &  0.51 & 5 (O)\\
  \#C- G \textit{[para]} & 5 & 0.09 & 200 & 10 & 10 & 0.58 & 5 (O)\\
  \hline
  Top- US \textit{[cross]} & 200 & 0.00 & 2 & 5 & 15 & 0.96 & 2\\
  Top- US \textit{[para]} & 100 & 0.09 & 2 & 5& 20 & 0.98 & 2\\
  \#C- US \textit{[cross]}  & 20 & 0.00 & 20 & 5 & 30 & 0.56 & 8 (O)\\
  \#C- US \textit{[para]} & 5 & 0.00 & 200 & 5 & 15 & 0.52 & 13 (O)\\
  \hline
  Top- I \textit{[cross]} & 200 & 0.00 & 2 & 5 & 10 & 0.99 & 2\\
  Top- I \textit{[para]} & 20 & 0.00 & 20 & 5 & 15 & 0.94 & 3 (O)\\
  \#C- I \textit{[cross]} & 5 & 0.09 & 2 & 5 & 10 & 0.91 & 4 (O)\\
  \#C- I \textit{[para]} & 20 & 0.00 & 20 & 5 & 10 & 0.91 & 4 (O)\\
  \end{tabular} 
\end{table}

\begin{figure}
\centering
\begin{subfigure}{.5\linewidth}
  \centering
\includegraphics[width=1\linewidth]{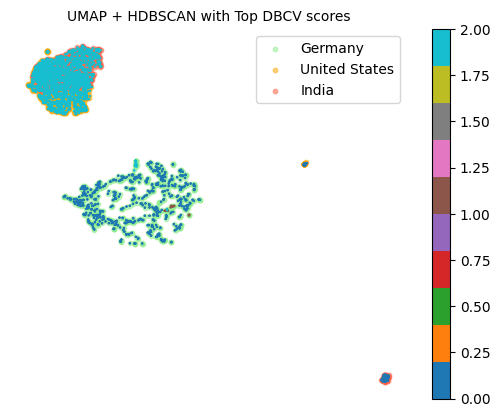}
  \caption{Embedding \textit{[cross]}}
  \label{fig:res_all_tsys}
\end{subfigure}%
\begin{subfigure}{.5\linewidth}
  \centering
  \includegraphics[width=1\linewidth]{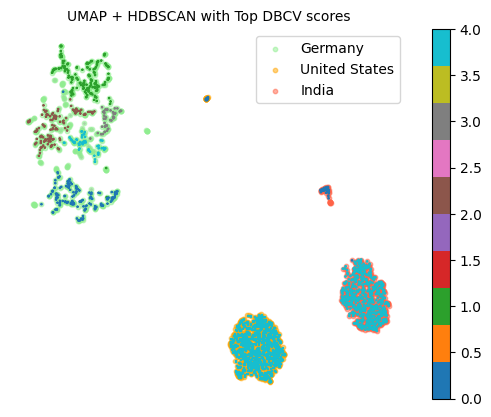}
  \caption{Embedding \textit{[para]}}
  \label{fig:res_all_para}
\end{subfigure}

\begin{subfigure}{.5\linewidth}
  \centering
\includegraphics[width=1\linewidth]{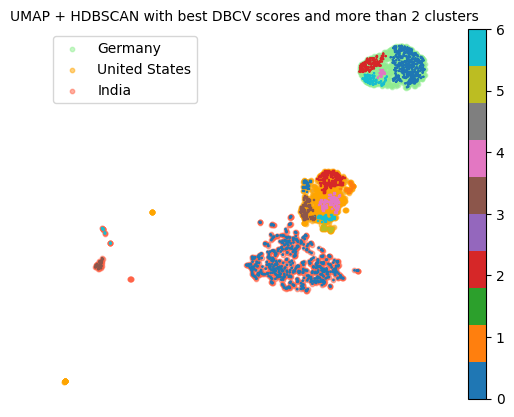}
  \caption{Embedding \textit{[cross]}}
  \label{fig:res_morecluster_tsys}
\end{subfigure}%
\begin{subfigure}{.5\linewidth}
  \centering
  \includegraphics[width=1\linewidth]{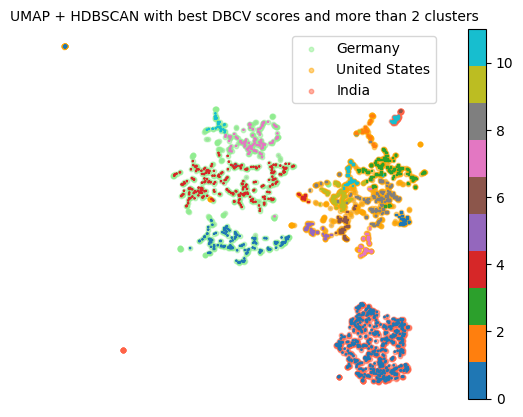}
  \caption{Embedding \textit{[para]}}
  \label{fig:res_morecluster_para}
\end{subfigure}
\caption{Clustering results of the fake news of Germany, United States and India using UMAP and HDBSCAN (Top and \#C)}
\Description{Clustering results in \textit{[cross]} and \textit{[para]} embedding using the hyperparameter settings for UMAP and HDBSCAN for Top and \#C.}
\label{fig:clusters_vis}
\end{figure}

In Figure \ref{fig:clusters_vis} the clustering results with the best hyperparameter configuration for each nation using \textit{[cross]} and \textit{[para]} as embedding are displayed. 
For Germany, the best DBCV value were obtained with setting n\_components (dimension) to 100. In the United States and India, the optimal value for n\_components was two. In Figure \ref{fig:clusters_vis}, for a better visualization, n\_components=2 was used. The clustering took place on n\_components= 100 for Germany and two for the United States and India.
In Figure \ref{fig:clusters_vis} each document is displayed as a point. It's colour shows, based on the colour bar on the right, the topic number. The frame of the points shows to which nation the document belongs to.
In Figure \ref{fig:res_all_tsys} one topic of India and United States were clustered in the same place, indicating a possible common topic. 
Using a different embedding (\textit{[para]}) these topics were clustered differently.
According to Figure \ref{fig:res_all_para} all detected topics of the three nations appeared to be different.

\begin{table}[th]
    \caption{Topic coherence results for 1) the best parameter configuration based on the DBCV score for UMAP and HDBSCAN (Top) and 2) best DBCV scores with more than 2 (3 with outlier) clusters (\#C). The results are given as the mean value of the topic results per nation. Nation: Germany [G], United States [US], India [I]}
  \label{tab:topic_coherence}
  \scriptsize
  \begin{tabular}{l|l|l|l|l|l|l}
  \hline
  Nation \cellcolor[gray]{0.9} &  \multicolumn{4}{c|}{ \cellcolor[gray]{0.9}} &  \multicolumn{2}{c}{CTC \cellcolor[gray]{0.9}} \\
  - Topic Extraction Method \cellcolor[gray]{0.9} & $c_v$ \cellcolor[gray]{0.9} & $u_{mass}$\cellcolor[gray]{0.9}& $c_{uci}$\cellcolor[gray]{0.9} & $c_{npmi}$ \cellcolor[gray]{0.9} & Intrusion\cellcolor[gray]{0.9} & Rating\cellcolor[gray]{0.9}\\
   \hline
   \hline
   \multicolumn{7}{l}{Parameter settings for UMAP and HDBSCAN based on top DBCV scores (Top): }\\
   \hline
   G \textit{[cross]} c-TF-IDF & 0.44	& \textbf{-5.24}	& \textbf{-6.92} & 	\textbf{-0.16}	&	0.78 &	\textbf{2.5}\\
   G \textit{[cross]} KeyBERT & 0.46 &	-15	& -10.45 & -0.37& 	\textbf{0.82} &	2.25 \\
   \textbf{G \textit{[para]} c-TF-IDF} & 0.44 &	-5.48	& -7.6	&-0.22	& 0.77 &2.0\\
   G \textit{[para]} KeyBERT & \textbf{0.48}&	-16.03&	-9.98	& -0.36	& 0.79 &	2.0\\
   \hline
   US \textit{[cross]} c-TF-IDF & 0.63 &	-1.09 &	-7.3	& -0.21	& 0.51 &	2.0\\
   US \textit{[cross]} KeyBERT & 0.43 &	-12.12 &	-8.15&	-0.29	&	0.74 &	2.0\\
   \textbf{US \textit{[para]} c-TF-IDF} & \textbf{0.64}&	\textbf{-1.01}	&-6.87 & \textbf{-0.2}	& \textbf{0.79}&	2.0\\
   US \textit{[para]} KeyBERT & 0.34	& -9.12	& \textbf{-6.64}	& -0.21	& 0.63 &	\textbf{3.0}\\
   \hline
   I \textit{[cross]} c-TF-IDF & 0.59 &-0.99 & -9.08&	-0.23	& 0.46 & 2.5\\
   I \textit{[cross]} KeyBERT & 0.42	& -12.74 & -7.16 & -0.22 & 0.66&	2.5\\
   I \textit{[para]} c-TF-IDF & \textbf{0.76} &	\textbf{-0.51}	& -7.34&	-0.13 &	0.48&	2.33\\
   \textbf{I \textit{[para]} KeyBERT} & 0.59	& -8.35 &	\textbf{-4.82} &	\textbf{-0.08}	& \textbf{0.7} &	\textbf{2.66} \\
   \hline
   \multicolumn{7}{l}{Parameter settings for UMAP and HDBSCAN with more than 2 clusters (\#C): }\\
   \hline
   G \textit{[cross]} c-TF-IDF & 0.36 & -12.2 & -8.93 & -0.29 & 0.64 & 2.0\\
   G \textit{[cross]} KeyBERT & 0.63 & -15.51 & -10.12 & -0.36 & 0.78 & \textbf{2.2}\\
   \textbf{G \textit{[para]} c-TF-IDF} & 0.44 & \textbf{-4.61} & \textbf{-5.64} & \textbf{-0.14} & 0.7 & \textbf{2.2}\\
   G \textit{[para]} KeyBERT & \textbf{0.67} & -17.66 & -9.74 & -0.35  & \textbf{0.79} & 2.0\\
   \hline
   US \textit{[cross]} c-TF-IDF & 0.56 & \textbf{-4.04} & \textbf{-4.25} & \textbf{-0.05} & 0.59 & \textbf{2.25}\\
   US \textit{[cross]} KeyBERT & 0.45 & -14.47 & -8.3 & -0.3 & 0.63 & 2.0\\
   \textbf{US \textit{[para]} c-TF-IDF} & \textbf{0.61} & -6.41 & -4.79 & -0.08 & 0.55 & 1.92\\
   US \textit{[para]} KeyBERT & 0.39 & -9.85 & -8.4 & -0.28 & \textbf{0.71} & 2.23\\
   \hline
   I \textit{[cross]} c-TF-IDF & 0.79 & -0.54 & -8.74 & -0.17 & \textbf{0.71} & 2.25\\
   I \textit{[cross]} KeyBERT & 0.54 & -7.16 & -8.96 & -0.3 & 0.58 & 2.0\\
   I \textit{[para]} c-TF-IDF & \textbf{0.81} & \textbf{-0.42} & -8.27 & -0.16 & 0.31 & 2.25\\
   \textbf{I \textit{[para]} KeyBERT} & 0.5 & -6.6 & \textbf{-5.19} & \textbf{-0.12} & 0.68 & \textbf{2.75}\\
  \end{tabular} 
\end{table}

\subsection{Topic Extraction}
Six different topic coherence measures were applied for evaluating the words extracted through c-TF-IDF and KeyBERT.
In Table \ref{tab:topic_coherence} the topic coherence results for first, the best parameter configuration of UMAP and HDBSCAN based on the DBCV score (Top) and secondly, for the parameter settings with the best DBCV score and more than two clusters (\#C) are given. 

In the setting of top DBCV scores, for Germany, the embedding \textit{[cross]} provided better results in terms of topic coherence. 
The topic coherence scores in the United States and India were better with the \textit{[para]} embedding. Therefore, for the results of the top DBCV setting, the \textit{[para]} embedding with the best performing topic representation (c-TF-IDF or KeyBERT) is given in Table \ref{tab:topics}, showing the final topics.

In the setting focusing on building more diverse clusters, contrary to the setting before, for Germany the \textit{[para]} embedding and for the United States the \textit{[cross]} embedding provided better topic coherence scores. 
However, as the topic coherence scores from India were also better in \textit{[para]} embedding, again, the results of \textit{[para]} embedding were used in the final topic evaluation.
For the topic representation, in both setting, for Germany and the United States c-TF-IDF provided better results, for India it was KeyBERT. The resulting keywords of these representations are shown in Table \ref{tab:topics}.

\begin{table*}[th]
    \caption{Topics extracted using 1) the best parameter configuration based on the DBCV score for UMAP and HDBSCAN and 2) the best DBCV scores with more than 2 (3 with outlier) clusters. Embedding: \textit{[para]}. Topic \# -1: Outlier; Underlined are the words that were extracted in both parameter settings of a nation. Topics that occur in multiple nations are written in bold letters.}
  \label{tab:topics}
  \scriptsize
  \begin{tabular}{l|l|l|l}
  \hline
   Nation\cellcolor[gray]{0.9} & \# \cellcolor[gray]{0.9} & Words describing the Topic \cellcolor[gray]{0.9}& Cluster Size (\# articles)\cellcolor[gray]{0.9}\\
   \hline
   \hline
   \multicolumn{4}{l}{Parameter settings for UMAP and HDBSCAN based on top DBCV scores (Top): }\\
   \hline
Germany (c-TF-IDF) & -1 & okt, herzstillstand, app, bewohner, senioren, volleyball, graphenoxid, spieler, herzinfarkt, 2021 & 95\\
 & 1 & \underline{masken}, \underline{maske}, \underline{tragen}, \underline{maskenpflicht}, \underline{mund}, \underline{merkel}, \underline{ffp2}, \underline{maßnahmen}, bundesregierung, michael & 160\\
& 2 & \underline{pcr}, \underline{kinder}, \underline{test}, \underline{gates}, \underline{israel}, \underline{pfizer}, mrna, \underline{geimpfte}, \underline{kindern}, \underline{tests} & 179\\
 & 3 & 5g, \underline{chlordioxid}, coronaviren, emf, viren, \textbf{china}, wissenschaftler, \textbf{wuhan}, grippe, coronavirus & 137\\
 & 4 & ade, \underline{antikörper}, dengue, neutralisierende, mers, verstärkung, risiko, cov, immunpathologie, ivermectin & 47\\
 & 5 & totenschein, zahlen, kliniken, \underline{impfungen}, weckruf, bestatter, totenscheine, todesfälle, impfpraxis, millionen & 57\\
\hline
United States (c-TF-IDF)& 1 & \underline{cookies}, \underline{partners}, \underline{agree}, \underline{free}, \underline{customize}, premiumeu, uphold, \underline{personalized}, \underline{analytics}, \underline{117} & 21\\
 & 2 &  way, \underline{prevent}, photo, institute, twitter, different, \underline{chinese}, \underline{fauci}, scientists, conspiracy & 774\\
\hline
India (KeyBERT)& -1 & factual, factly, fact, journalism, news, sources, data, evidence, india, information & 9\\
 & 1 & \underline{factual}, \underline{factly}, facts, \underline{fact}, \underline{sources}, \underline{news}, \underline{debunked}, \underline{source},  \underline{journalism}, \underline{news18} & 49\\
 & 2 & \underline{newsflare}, \underline{covid2019india}, \underline{fakenews}, \underline{coronavirus}, \underline{coronavirusoutbreakindia}, \underline{fakenewsalert}, \underline{berita}, \underline{virall}, \underline{coronavirus}, \underline{newsy} & 595\\
\hline
\multicolumn{4}{l}{Parameter settings for UMAP and HDBSCAN with more than 2 clusters (\#C): }\\
\hline
Germany (c-TF-IDF) & -1 & protein, app, emf, \textbf{5g}, spike, exposition, graphenoxid, kreislauf, gehirn, mrna & 100\\
& 1 & \textbf{\underline{masken}}, \textbf{\underline{maske}}, \textbf{\underline{tragen}}, \underline{maskenpflicht}, \underline{mund}, \underline{ffp2}, \textbf{\underline{merkel}}, \underline{maßnahmen}, bevölkerung, freiheit & 156\\
& 2 & patienten, coronaviren, \underline{antikörper}, \underline{chlordioxid}, cov, senioren, schwere, \textbf{\underline{impfungen}}, schweren, 2021 & 275\\
& 3 & \underline{gates}, \underline{israel}, nebenwirkungen, \textbf{\underline{kinder}}, \underline{pfizer}, \textbf{\underline{geimpften}}, \textbf{geimpfte}, biontech, alter, \underline{kindern} & 112\\
& 4 & \underline{pcr}, \textbf{\underline{test}}, drosten, nürnberger, \underline{tests}, kodex, spezifität, positiv, falsch, sensitivität & 32\\
\hline
United States (c-TF-IDF) & -1 & pope, zoom, beach, election, goodwill, wisconsin, vatican, children, county, 5g & 106\\
& 1 & \underline{cookies}, \underline{partners}, \underline{agree}, \underline{free}, premiumeu, \underline{customize}, uphold, \underline{personalized}, \underline{analytics}, \underline{117} & 20\\
& 2 & \textbf{mask}, \textbf{wearing}, face, cloth, wear, \textbf{masks}, coverings, n95, particles, air & 36\\
& 3 & vitamin, uv, water, sanitizer, hand, kill, treat, cure, \underline{prevent}, drinking  & 48\\
& 4 & ncov, \textbf{\underline{chinese}}, lab, animal, \textbf{wuhan}, coronaviruses, animals, cov, humans, syndrome  & 156\\
& 5 & \underline{fauci}, chloroquine, hydroxychloroquine, novartis, anthony, 2005, microchip, drug, remdesivir, corsi  & 19\\
& 6 & \textbf{whitmer}, \textbf{michigan}, \textbf{governors}, order, governor, gov, gretchen, stay, orders, flags  & 46\\
& 7 & \textbf{biden}, \textbf{obama}, \textbf{trumps}, h1n1, donald, labs, administration, \textbf{tests}, white, test  & 59\\
& 8 & stimulus, senate, pelosi, tax, relief, package, aid, trillion, money, debt  & 46\\
& 9 & \textbf{vaccines}, nursing, influenza, rna, \textbf{vaccination}, ceo, ventilators, smoking, foundation, cuomo  & 134\\
& 10 & boxes, empty, pence, satire, kimmel, var, hanks, blair, hoax, edited  & 81\\
& 11 & door, walmart, stockton, crab, shave, schools, graphic, police, beards, ammunition  & 22\\
& 12 & mikovits, \textbf{5g}, encryption, ayyadurai, earn, end, exploitation, \textbf{child}, tech, ventura  & 22\\
\hline
India (KeyBERT) & -1 & factual, factly, fact, sources, journalism, india, verify, media, reports, news & 10\\
& 1 & \underline{newsflare}, \underline{covid2019india}, \underline{fakenews}, \underline{coronavirus}, \underline{coronavirusoutbreakindia}, \underline{fakenewsalert}, \underline{berita}, \underline{virall}, \underline{coronavirus}, \underline{newsy} & 595\\
& 2 & \underline{factual}, \underline{factly}, \underline{fact}, \underline{news}, journalism, \underline{sources}, video, india, data, youtube & 14\\
& 3 & factual, factly, facts, fact, news, \underline{sources}, \underline{debunked}, \underline{journalism}, \underline{news18}, source & 34\\
  \end{tabular} 
\end{table*}

\section{Discussion}\label{sec:discssion}
We attempted to analyse the applicability of BERTopic on a practical use case focusing on whether fake news about the Covid-19 pandemic varied between different countries.
According to the clustering in Figure \ref{fig:res_all_tsys}, the fake news of the United States and India were close to each other. However, using a different embedding, e.g. Figure \ref{fig:res_all_para}, all nations were clustered differently. Whereby, it is important to keep in mind that the data points were clustered in a higher dimensional space and then transformed into two dimensions for visualization purposes.
Therefore, to identify commonalities or differences in fake news it is important to compare the final topics rather than the clustered visualized.

\subsection{Use Case: Topic Modeling of Fake News}
Using topic coherence, the best topic extraction method per nation was found and the resulting top 10 words per topic were displayed in Table \ref{tab:topics}.
In the first setting, using the best performing parameter setting for UMAP and HDBSCAN, in the fake news of Germany five topics were formed and can be identified clearly. 
For example, the first, described with words such as `masks', `mask requirement' and `ffp2' presumably is about the obligation to wear a mask.
In the same setting, the topic describing words of the United States and India were not that clearly interpretable. Probably also only two topics and an outlier for India were recognised.
Both, for the news articles from the United States and India, one big and a small topic in terms of number of articles, were built.
In the diverse clustering approach, however, we were able to identify better interpretable results.

In the parameter setting that focuses on building more clusters, for Germany, only four topics were built but again easily interpretable (see Table \ref{tab:topics}).
The topics in the fake news from Germany are similar in both settings.
In the parameter setting that focuses on building more clusters, particularly in the United States, one can see that more diverse and detailed topics have been created. These topics can now be easily identified.
For topic 1, again, the same topic was built as in the first setting, now with 20 articles (before: 21 articles). 
However, the other topic has now been divided into 11 smaller topics instead of being one large topic.
The main topic with 156 articles (topic 4) seems to be about the presumed origin in China, Wuhan and Chinese labs, similar to topic 3 in Germany in the first setting.
In India, the same topics as in the first setting were built, which were again not interpretable.
This could also be due to the possibility that there are really only one or two topics in the Indian news.
The German data set is more diverse because the news was collected over a longer period of time.
In the US, proper configuration of settings aided in better topic recognition.

\subsection{Analysis of Evaluation Results}
When focusing on the evaluation results in Table \ref{tab:hyperparameter_DBCV} and \ref{tab:topic_coherence} certain numbers stand out.
First, the DBCV score of Germany using the \textit{[cross]} embedding led in both settings to lower DBCV scores (e.g. Top setting: \textit{[cross]}: 0.57; \textit{[para]}: 0.6) as the \textit{[para]} embedding. As \textit{[cross]} was included in this evaluation for its better performance on German data, this was not expected.
It can be assumed that since \textit{[para]} was trained on a bigger and more diverse data corpora, a better representation was generated.
In parallel, the DBCV score for Germany were lower in all settings, compared to the other two nations, in the United States and India reaching scores of up to 0.99. 
We assume that for English, as one of the main languages of the world and in research, the models are trained to generate more suitable embeddings for English.
Contrarily, when viewing the topic coherence scores, in Germany, \textit{[cross]} provided better results in terms of topic coherence measures in the first (Top) setting. But, in the second setting using \textit{[para]} provided superior results.

The different choices of the embedding resulted in different numbers of clusters, being for example four in the \textit{[cross]} embedding and six in the \textit{[para]} embedding (Top setting), which, as we saw in this analysis, has a significant impact on the resulting topics.
When viewing the topics generated, one can see that with a lower number of clusters, in the United States and India the topics are partially not recognisable. For this purpose, to generate more diverse clusters the settings with the best DBCV scores for more than 2 (3 with outlier) clusters were additionally observed, which helped us to identify common topics between Germany and the United States leading to the best result.
In India and the United States all top five hyperparameter configurations built two topics (or three and one outlier topic).
Only as viewing lower DBCV scores, parameter settings with higher cluster numbers were formed. 
We assume that when choosing a hyperparameter configuration with a lower DBCV score but even more numbers of clusters, also the topics in the Indian dataset can be identified.
We could observe that the DBCV score does not naturally correspond to the quality of the end results.
Our results show that it might be important to vary the hyperparameter settings and also to consider settings outside the best evaluation scores when using topic modeling to analyse a dataset.
Moreover, from our subjective analysis we could observe that in both settings, for Germany and the United States c-TF-IDF provided the best topic representation, for India it was KeyBERT, although the $CTC_Rating$ were better for KeyBERT in the United States dataset. Therefore, we can again observe similar results as with DBCV, that topic coherence do not directly represent the actual quality of final topic words.

\section{Conclusion}\label{conclusion}
This work applied and analysed state-of-the-art sentence embeddings and clustering on a multilingual text corpora to identify the topics of fake news of different nations in two languages.
We successfully extracted the topics of fake news about the Covid-19 pandemic from two countries in two different languages and found similarities in the fake news of Germany and the United States.
We have also shown how important the parameter selection for HDBSCAN and UMAP is in order to obtain interpretable topics and that the selection of parameters cannot depend solely on the evaluation metrics.
The most interpretable results were not achieved with the best DBCV values for hyperparameter configurations of HDBSCAN and UMAP but with their combination with an increased cluster number.
In this work, the similarities between the topics were determined subjectively based on the words extracted. An approach for future work is an objective evaluation of the similarity of the words using cosine similarity.
The topic model used here assigns each article to only one topic, but several topics can be included in a single article which could be an approach that can be used to further analyse the India datasets.
Clustering techniques that take into account a data point as belonging to multiple clusters can be used in future research.

\begin{acks}
This research work was supported by the National Research Center for Applied Cybersecurity ATHENE.
\end{acks}

\bibliographystyle{ACM-Reference-Format}
\bibliography{main}


\end{document}